\documentclass[lettersize,journal]{IEEEtran}
\usepackage{amsmath,amsfonts}
\usepackage{algorithmic}
\usepackage{algorithm}
\usepackage{array}
\usepackage[caption=false,font=normalsize,labelfont=sf,textfont=sf]{subfig}
\usepackage{textcomp}
\usepackage{stfloats}
\usepackage{url}
\usepackage{verbatim}
\usepackage{graphicx}
\usepackage{cite}
\usepackage{booktabs}
\usepackage{multirow}
\usepackage[table]{xcolor}
\usepackage{adjustbox}
\usepackage{pifont}
\usepackage{makecell}
\begin{document}

\title{Look Back, Correct Forward: Error Feedback Calibration for Time Series Forecasting}
\author{
  Haonan Yang,
  Yulin He\textsuperscript{*},
  Zhuo Li,
  Jing Zhao,
  and Huibin Tan
  \thanks{\textsuperscript{*} Corresponding author. E-mail: heyulin@nudt.edu.cn}
  \thanks{All authors are with the College of Computer Science and Technology, National University of Defense Technology, Changsha 410073, China.}
}



\maketitle

\begin{abstract}
Time series forecasting models are increasingly deployed in continuously evolving environments. After a forecast is issued, future observations gradually arrive and reveal how the model has failed. Existing methods mainly treat such observations as supervision for parameter updates. However, it remains unclear whether verified forecasting errors can themselves provide temporally available feedback for correcting future predictions. To address this, we introduce Error Feedback Calibration (EFC), a fully test-time framework for deterministic post-hoc correction of time series forecasting. EFC constructs a Causal Error Diagonal (CED) from revealed errors of overlapping historical forecasts, providing a fresh view of the frozen forecaster's recent failure behavior. Then Cumulative Error Transport (CET) maps this error state to future residuals using cumulative estimation, while temporal out-of-sample validation controls the correction strength. Temporal Residual Learning (TRL) further predicts only the residual structure left after CET from recent CED trajectories, rather than relearning the entire forecasting error. Extensive experiments across forecasting backbones and datasets show that EFC broadly improves forecasting performance, with temporally out-of-sample reliability estimates used to regulate correction strength.
\end{abstract}

\begin{IEEEkeywords}
Time series forecasting, test-time adaptation, error feedback, post-hoc forecast correction.
\end{IEEEkeywords}

\section{Introduction}
\IEEEPARstart{T}{ime} series forecasting (TSF) has advanced rapidly with deep learning \cite{TSL2024,yang2025}, leading to increasingly accurate models across a wide range of forecasting tasks \cite{Kaifeng2023,Chenjuan2020}. Yet strong average performance \cite{TimeMixer2024,TimesNet2023} does not guarantee stable behavior after deployment. Real world time series are often non-stationary \cite{Zhiding2023}, and distribution shifts \cite{Autoformer2021} or regime changes \cite{yang2026} can weaken patterns learned from historical data \cite{Qiuz2024}. As the test stream evolves, a previously accurate forecaster may develop systematic errors, such as persistent bias and delayed responses to changing trends \cite{liang2026the}. A model that performs well under offline evaluation may therefore degrade during deployment \cite{ning2025}. This motivates the need for forecast adaptation or correction as the underlying data evolves \cite{dytta2026}.

To address such performance degradation during deployment, recent studies have explored post-hoc forecast revision \cite{PIR2025,kudrat2025,FreDF2025} and test-time adaptation \cite{medeiros2025,kim2025,im2026cosa}. Post-hoc methods revise the forecasts after they are generated \cite{PIR2025,dbloss2025,zhao2026tackling}, while test-time adaptation typically updates model parameters \cite{medeiros2025} or lightweight adapter \cite{kim2025} as the test stream evolves. Forecasting has an important temporal property: historical predictions are progressively verified as their outcomes become available \cite{wang2026quadratic}. Existing methods use this information to decide which forecasts should be revised or which model components should be updated \cite{Zezhi2025}. The revealed forecasting error, however, provides information beyond the new observation itself. A new observation describes what happened in the data stream, whereas the error of an earlier forecast reveals how the forecaster's prediction deviated from the observed outcome \cite{FreDF2025}. This suggests a different perspective: revealed forecasting errors can be treated not only as signals for adaptation, but also as observations of how the forecaster has recently failed \cite{cao2026enhancing}.

\begin{figure*}[!t]
    \centering
    \includegraphics[width=\linewidth]{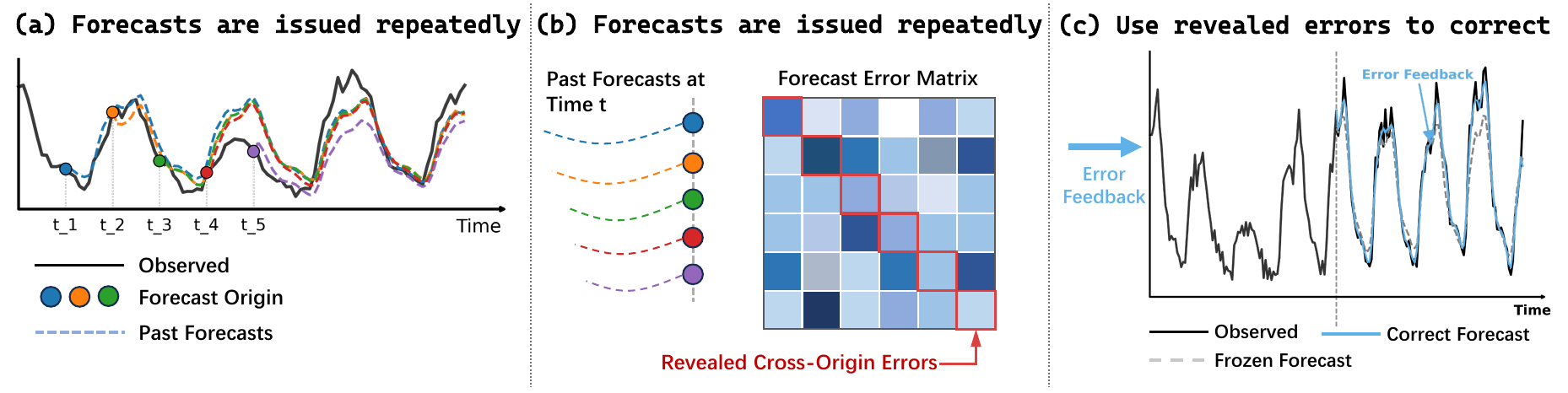}
    \caption{Forecast errors as causal feedback. Historical point forecasts issued at different origins overlap at the same target time. Once its observation becomes available, their errors at different lead times are revealed. Rather than discarding these verified errors after evaluation, error feedback correction uses them to inform subsequent forecasts from a frozen forecaster.}
    \label{fig:introd}
\end{figure*}

This perspective raises a central question: can forecasting errors that have already occurred and been verified during deployment serve as temporally valid feedback for correcting future errors of a frozen forecaster? Fig. \ref{fig:introd} illustrates how overlapping forecasts provide verified error feedback during deployment. Because multi-step forecasts are repeatedly issued from overlapping origins, a newly arrived observation can simultaneously verify several predictions made earlier and at different lead times \cite{TimesNet2023}. These revealed errors are already available before the next forecast is issued, providing temporally valid feedback on the forecaster's recent prediction errors. Turning revealed errors into useful corrections, however, is not straightforward \cite{qiao2026it}. First, the complete error of the current forecast is unavailable until its full prediction horizon has elapsed. Relying only on the latest fully matured error therefore introduces time delay. The first challenge is to organize these revealed cross-origin errors into a timely representation of the current forecasting error state. Second, recent errors cannot be assumed to persist unchanged into the future. Their relationship with future residuals can be noisy and horizon dependent, and directly reusing these errors may lead to over correction. The second challenge is therefore to estimate this relationship and assess the reliability of the resulting correction during deployment. Third, it remains unclear whether residual left after the initial correction contains temporal structure that can be predicted from recent error histories. A practical error feedback method must therefore observe the current error state in time, transport it reliably to future residuals, and identify what remains predictable afterwards.

To address these challenges, we propose Error Feedback Calibration (EFC), a fully test-time framework that uses already revealed forecasting errors to correct future predictions of a frozen forecaster. EFC first introduces the Causal Error Diagonal (CED), which organizes verified errors for the same target time from different historical forecast origins and lead times. CED provides a timely cross-origin view of the forecaster's recent error behavior without waiting for the current forecast to mature. Based on this representation, Cumulative Error Transport (CET) estimates a stable mapping from revealed error states to future residuals using fully matured historical error pairs, while temporally out-of-sample reliability estimates control the correction strength. Temporal Residual Learning (TRL) then models only the residual left after CET from recent CED trajectories, rather than relearning the entire forecasting error. A separate temporally out-of-sample reliability estimate controls the strength of the TRL correction. Together, EFC uses verified historical errors to correct future forecasts while keeping the backbone frozen and excluding future observations unavailable at the current forecast origin. Our contributions are summarized as follows:

\begin{itemize}
    \item We formulate verified forecasting errors as causal feedback and introduce the Causal Error Diagonal (CED) to organize revealed errors across historical forecast origins and lead times into a temporally available representation.
    \item We develop Cumulative Error Transport (CET) to estimate a stable mapping from revealed error states to future residuals using fully matured historical error pairs. CET further separates estimation from deployment through temporally out-of-sample reliability, so that an estimated correction is applied only when supported by independent historical evidence.
    \item We also introduce Temporal Residual Learning (TRL) to predict the residual remaining after the CET correction actually issued at each historical origin, rather than relearning the full forecasting error. TRL uses recent CED trajectories to capture remaining temporal predictability and evaluates its incremental value with temporally out-of-sample evidence.
    \item Extensive experiments across diverse real-world datasets and state-of-the-art backbones show that EFC improves frozen forecasting models across a broad range of settings with modest absolute inference-time overhead for most evaluated backbones.
\end{itemize}

\section{Related Work}
\subsection{Time Series Forecasting Models}
Deep learning has become a major approach for time series forecasting, with increasingly diverse architectures designed to capture complex temporal patterns \cite{qiu2025c, li2025tsfm}. MLP-based models demonstrate that strong forecasting performance can be achieved with simple linear layers \cite{NBEATS2019,FreTS2023}. CNN-based approaches capture local and multi-scale temporal patterns through hierarchical feature extraction \cite{TCN2018,SCINet2022,MiCN2023}. Transformer-based models further extend this line of research by modeling long-range dependencies and developing representations tailored to time series \cite{Informer2021,iTransformer2023,pathformer2024}. Representative designs incorporate decomposition \cite{Autoformer2021}, patch-based tokenization \cite{PatchTST2023}, and variate-wise attention \cite{iTransformer2023} to improve long-term forecasting. More recently, time series foundation models have moved beyond dataset-specific training \cite{MOIARI2024,Timer2024,TimeMoE2025}. By pretraining on large and heterogeneous time series data, these models aim to provide transferable or zero-shot forecasting across domains, horizons, and temporal resolutions \cite{liu2026sundial,jiang2026timexl,chen2025visionts}. 

Existing forecasting architectures primarily improve mappings from historical observations to future values \cite{qiao2026it,Ailing2023}. A stronger forecasting backbone may reduce average errors, but its error patterns can still change as the data evolve during deployment. More importantly, once forecasts have been issued and their errors are later revealed, backbone design alone does not specify how this newly verified information should be reused. Our work addresses this complementary problem. EFC keeps the forecasting backbone frozen and asks whether errors revealed during deployment can be reused as causal feedback to correct subsequent forecasts.

\subsection{Post-hoc Forecast Revision}
Even with a strong forecasting backbone, individual predictions can still exhibit large errors \cite{dbloss2025}. This has motivated post-hoc forecast revision, which seeks to improve a forecast after the base prediction has already been produced \cite{PIR2025,kudrat2025,kayaalp2026}. Recent work has explored identifying unreliable forecasting instances and revising them with local or global contextual information \cite{PIR2025}. Other post-processing approaches refine the input or output interface of a deployed forecaster and learn lightweight corrections to its original predictions \cite{liang2026the, kudrat2025}. These studies show that forecast generation need not be the final step of the forecasting pipeline and that an issued prediction can still be improved through model-agnostic post-processing.

Recent studies have explored error-driven corrections for frozen forecasters. ResCAL \cite{kimrescl} exploits historical forecasting residuals and graph signals to estimate future errors for real-time traffic forecast calibration. STEPS \cite{liu2026steps} propagates observed prefix errors within an issued forecast, while FreshCast \cite{he2026retrieval} uses streaming retrieval memory to construct calibrated corrections. EPOC \cite{fujimoto2026epoc} compresses historical residuals using DCT coefficients and endpoint information for online ridge correction. CRC \cite{xie2026cau} instead employs explicit safeguards against harmful residual revisions. In contrast, EFC constructs a cross-origin, multi-lead error state from verified historical forecasts before issuing each new prediction and learns its relationship with future forecasting residuals.

\subsection{Test-Time Adaptation for Time Series Forecasting}
Test-time adaptation (TTA) has recently been introduced into time series forecasting to address distribution shifts that arise after deployment \cite{dytta2026}. Unlike many conventional TTA settings, time series forecasting progressively reveals target observations that can provide supervision for online adaptation \cite{medeiros2025}. TAFAS \cite{kim2025} uses partially observed ground truth and gated calibration modules to adapt a frozen forecasting backbone. Subsequent work has focused on making this adaptation more efficient and more targeted. PETSA \cite{medeiros2025} updates only lightweight low-rank calibration modules at the input and output sides of the forecaster, reducing the cost of test-time adaptation. COSA \cite{im2026cosa} further moves adaptation into the output space, where a lightweight adapter directly corrects predictions using the base forecast and statistics from recently observed ground truth. TEFL \cite{huang2026tefl} further incorporates observable residuals from rolling forecasts during both training and evaluation, using a lightweight low-rank adapter and a two-stage training procedure that jointly optimizes the forecasting backbone and error module. These studies exploit revealed observations or forecasting residuals through different forms of test-time calibration, output-space correction, and residual-aware training.

EFC operates in the same deployment setting as these approaches, but differs in how newly available feedback is represented and used. While existing methods leverage revealed observations or forecasting residuals for adaptation and calibration \cite{wang2026principle}, EFC organizes verified errors from overlapping historical forecast origins and lead times into CED. This representation provides a timely view of the forecaster's recent error state as each observation arrives, whereas CET and TRL learn correction relationships only from fully matured historical forecasting errors. By separating immediate error sensing from delayed correction learning, EFC complements existing TSF-TTA approaches while keeping the forecasting backbone frozen.

\begin{figure*}[!t]
    \centering
    \includegraphics[width=0.9\linewidth]{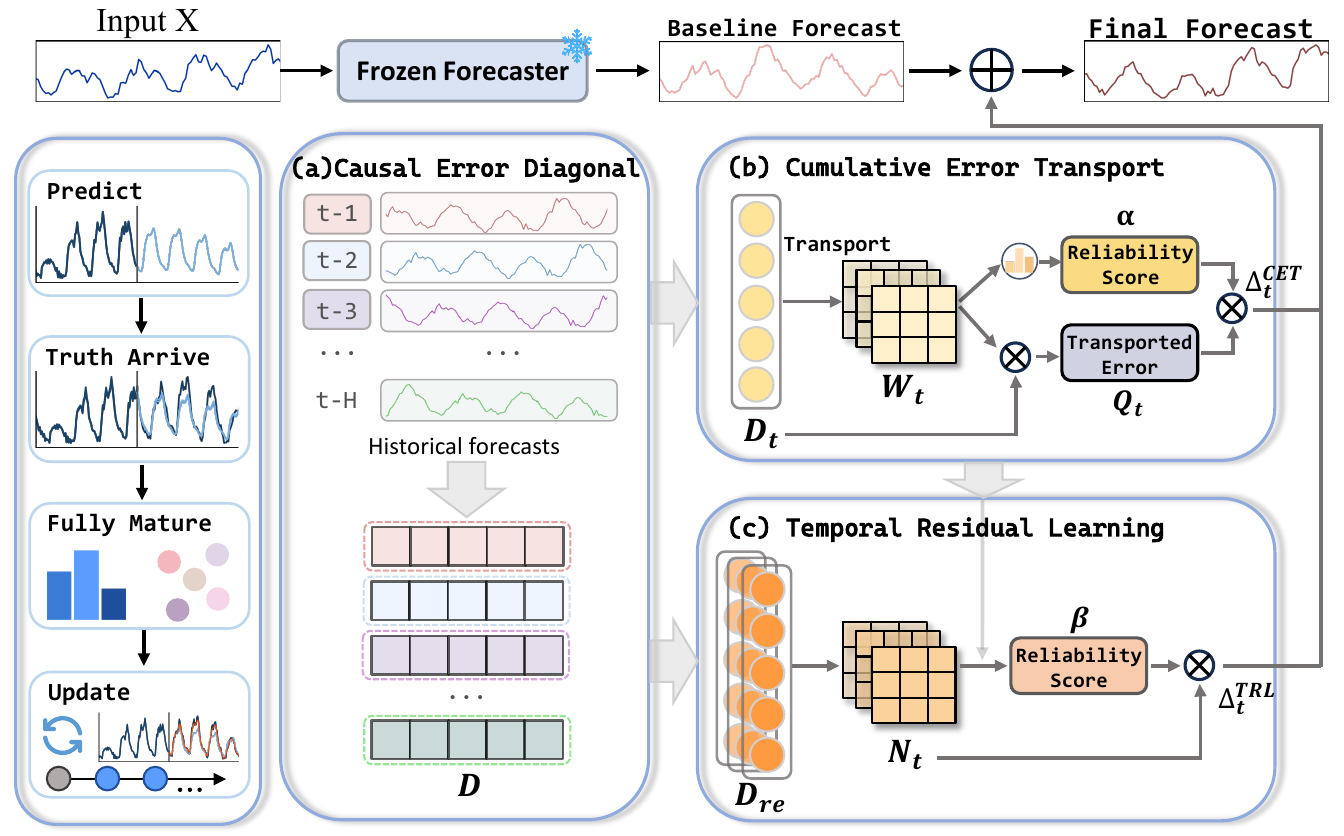}
    \caption{Overview of Error Feedback Calibration (EFC). Upon observing the current target, CED collects verified errors from overlapping historical forecasts at different origins and lead times. CET learns cumulative error transport from fully matured historical samples, while TRL models the remaining CET-deployment residual using recent CED states.}
    \label{fig:method}
\end{figure*}

\section{Method}
Time series forecasting aims to predict future values ${Y} = \{y_{L+1}, ..., y_{L+H}\}\in\mathbb{R}^{H\times{C}}$ from historical observations ${X} = \{x_1, ..., x_L\}\in\mathbb{R}^{L\times{C}}$, where $H$ is the prediction horizon, $L$ is the history horizon, and $C$ is the number of variables.

\subsection{Structure Overview}
As illustrated in Fig. \ref{fig:method}, EFC uses the errors revealed during deployment to calibrate the predictions of a frozen forecasting backbone. EFC consists of a shared error representation and two correction components. First, the Causal Error Diagonal (CED) provides a representation of the forecaster's recent error behavior. Based on CED, Cumulative Error Transport (CET) estimates the error structure that can be reliably transported to future residuals. Temporal Residual Learning (TRL) further models the residual structure that remains predictable after the CET correction by using recent CED trajectories.

\subsection{Causal Error Diagonal (CED)}
The complete forecasting error $E_t=Y_t-\hat{Y}_t$ is unavailable when the forecast at origin $t$ is issued, as its future observations have not yet arrived. Using the latest fully matured error $E_{t-H}$ therefore introduces an $H$-step delay. However, a newly arrived observation $y_t$ can simultaneously verify multiple historical forecasts. Specifically, a forecast issued at origin $t-h$ contains an $h$-step prediction of $y_t$. These already revealed errors provide timely information about the frozen forecaster's recent behavior.

To organize this information, we introduce the Causal Error Diagonal (CED). For clarity, we describe the construction for each variable and omit the channel index. Let $\hat y_{t-h}^{(h)}$ denote the $h$-step prediction issued at origin $t-h$. For each variable, CED is defined as: 

\begin{equation}
D_t[h]=y_t-\hat y_{t-h}^{(h)},
\qquad h=1,\ldots,H.
\end{equation}

Equivalently, $D_t=[e_{t-1,1},e_{t-2,2},\ldots,e_{t-H,H}]\in\mathbb R^{1\times H},$ where $e_{i,h}=y_{i+h}-\hat y_i^{(h)}$. Each entry of $D_t$ corresponds to the same observed target $y_t$, but originates from a different historical forecast and lead time. These errors therefore form a diagonal in the origin-horizon error grid. CED thus provides a cross-origin, multi-lead observation of the forecaster's current error state.

Constructing $D_t$ requires only the current observation $y_t$ and the preceding $H$ issued forecasts. A bounded buffer of these $H$-step forecasts is sufficient, requiring $O(CH^2)$ storage independent of stream length. Here causal refers to temporal information availability rather than causal identification.


\subsection{Cumulative Error Transport (CET)}
While CED captures the forecaster's currently observable error state, it does not directly specify the residual of the current $H$-step forecast. We introduce Cumulative Error Transport (CET) to estimate the relationship between previously verified error states and their subsequent forecasting residuals. Let $\mathcal M_t={\{i|i+H\le t\}}$ denote the set of historical forecast origins whose complete errors are available at time $t$. For each variable, CET estimates a linear transport matrix $W_t\in\mathbb R^{H\times H}$ from the matured pairs ${(D_i,E_i)}_{i\in\mathcal M_t}$, where $D_i,E_i\in\mathbb R^{1\times H}$. The channel index is omitted for clarity. The cumulative estimator is computed as follows:

\begin{equation}
W_t=S_t^{-1}T_t,
\end{equation}
\begin{equation}
\begin{aligned}
S_t&=\lambda I+\sum_{i\in\mathcal M_t}D_i^\top D_i,
T_t&=\sum_{i\in\mathcal M_t}D_i^\top E_i,
\end{aligned}
\end{equation}

where $\lambda>0$ is a ridge regularization coefficient that initializes the cumulative statistic as $S_0=\lambda I$. CET maintains the inverse regularized Gram matrix $P_t=S_t^{-1}$ and the transport matrix $W_t$, updating both through per-variable rank-one RLS whenever a historical error pair matures. This avoids repeated dense matrix solves and requires $O(CH^2)$ operations per matured multivariate pair. The resulting candidate correction is $Q_t=D_tW_t$.

To control the strength of the transported correction, CET estimates its reliability using temporally out-of-sample validation evidence. Let $\mathcal V_t^\alpha\subseteq\mathcal M_t$ denote a window of fully matured validation origins. We use a saved transport estimator $W_v$ fitted only on historical origins preceding this window, satisfying $\operatorname{fit}{\max}(W_v)<\min\mathcal V_t^\alpha$. We evaluate its predictions $Q{i|v}=D_iW_v$ against the corresponding matured errors and obtain $\alpha$ as follows:

\begin{equation}
    \alpha_t=
\operatorname{clip}{[0,1]}
\left(
\frac{
\sum{i\in\mathcal V_t^\alpha}
\langle Q_{i|v},E_i\rangle
}{
\sum_{i\in\mathcal V_t^\alpha}
|Q_{i|v}|_2^2+\epsilon
}
\right).
\label{eq:1}
\end{equation}

The reliability coefficient in (\ref{eq:1}) can be interpreted as the solution to a regularized least-squares problem on temporally out-of-sample, minimizing $\sum_{i\in\mathcal V_t^\alpha}\|E_i-\alpha Q_{i|v}\|_F^2+\epsilon \alpha^2$ over $\alpha\in[0,1]$. The constraint suppresses corrections with nonpositive aggregate alignment and prevents amplification of the candidate correction. Although CET estimates a separate transport matrix for each variable, the inner products and squared norms in (\ref{eq:1}) are aggregated across variables and forecast steps to obtain a shared reliability coefficient. This does not allow variable- or lead-specific correction strengths, but reduces the number of scaling coefficients to estimate. This coefficient scales the CET correction according to its agreement with historical errors evaluated out of sample. We further smooth it using an exponential moving average:

\begin{equation}
\bar\alpha_t=\rho\bar\alpha_{t-1}
+
(1-\rho)\alpha_t,
\qquad \rho=0.95.
\end{equation}

The EMA is updated only at scheduled reliability recomputation checkpoints, while the previous smoothed coefficient is retained between checkpoints. The same update convention applies to TRL reliability. Then the deployed CET correction is computed as:

\begin{equation}
    \Delta_t^{\mathrm{CET}}=
    \bar\alpha_tD_tW_t.
\end{equation}

During initialization, the CET correction remains inactive until a valid historical estimator and out-of-sample reliability estimate are available. The correction actually issued at each forecast origin is saved so that TRL can subsequently learn the residual remaining after CET.

\subsection{Temporal Residual Learning (TRL)}
CET estimates a cumulative mapping from the current CED to future forecasting errors. However, the residual remaining after CET may still contain temporal patterns that can be learned from recent error feedback. We introduce Temporal Residual Learning (TRL) to predict the residual remaining after the deployed CET correction. For each historical forecast origin $i$, we define this post-CET residual as $R_i^{\mathrm{base}}$:

\begin{equation}
    R_i^{\mathrm{base}} = E_i-\Delta_i^{\mathrm{CET}} = E_i-\bar{\alpha}_iD_iW_i,
\end{equation}

where $\Delta_i^{\mathrm{CET}}$ is the correction actually issued at origin $i$ and saved at that time. Once the complete forecasting error $E_i$ becomes available, $R_i^{\mathrm{base}}$ provides the learning target for TRL. This allows TRL to focus on the error remaining after the deployed CET correction.

To capture its temporal structure, TRL takes a sequence of recent CED states $\mathcal D_t = [D_{t-R+1},\ldots,D_t]$ as input and produces a residual prediction:

\begin{equation}
    N_t=f_{\theta_t}(\mathcal D_t).
\end{equation}

The function $f_{\theta_t}$ consists of an input projection with dimension $z$, a GRU with hidden dimension $d_{\mathrm{hid}}$, and an output projection. It is trained on fully matured historical sequence-residual pairs $(\mathcal D_i,R_i^{\mathrm{base}})$ using the Huber loss:

\begin{equation}
    \mathcal L_i^{\mathrm{TRL}} = \operatorname{Huber} \left( f_\theta(\mathcal D_i), R_i^{\mathrm{base}} \right).
\end{equation}

TRL thus learns to predict the residuals remaining after deployed CET corrections from recent CED trajectories.

We further estimate the incremental reliability of TRL using temporally out-of-sample evidence. Let $\mathcal V_t^\beta$ be a window of fully matured validation origins and $\theta_v$ a saved model satisfying $\operatorname{train}_{\max}(\theta_v) < \min \mathcal V_t^\beta$. With $N_{i|v} = f_{\theta_v}(\mathcal D_i)$, the reliability coefficient is defined as:

\begin{equation}
    \beta_t = \operatorname{clip}_{[0,1]} \left( \frac{ \sum_{i\in\mathcal V_t^\beta} \langle N_{i|v},R_i^{\mathrm{base}}\rangle }{ \sum_{i\in\mathcal V_t^\beta} \|N_{i|v}\|_2^2+\epsilon } \right).
    \label{eq:2}
\end{equation}

The same constrained least-squares interpretation applies to $\beta_t$ in (\ref{eq:2}), with $R_i^{\mathrm{base}}$ as the target and $N_{i|v}$ as the candidate residual correction. As in CET, the inner products and squared norms are aggregated across variables and forecast steps to obtain a shared coefficient. After exponential moving average smoothing, $\bar{\beta}_t = \rho\bar{\beta}_{t-1} + (1-\rho)\beta_t$, the deployed TRL correction is:

\begin{equation}
    \Delta_t^{\mathrm{TRL}} = \bar{\beta}_t f_{\theta_t}(\mathcal D_t).
\end{equation}

When no eligible out-of-sample estimator is available, TRL also remains inactive. The final EFC forecast combines the frozen backbone prediction with the CET and TRL corrections:
\begin{equation}
    \begin{aligned} \hat Y_t^{\mathrm{EFC}} ={}&\hat Y_t+\bar\alpha_tD_tW_t +\bar\beta_t f_{\theta_t}(\mathcal D_t). \end{aligned}
\end{equation}

\subsection{Fully Test-Time Deployment}
EFC operates sequentially as new observations arrive, while keeping the forecasting backbone frozen. At each time step $t$, the newly observed $y_t$ makes historical forecasts eligible for learning once their complete prediction horizons have matured. CET updates its RLS state whenever a historical error pair fully matures. Reliability recomputation and TRL optimization occur separately when their eligibility and checkpoint conditions are satisfied. Next, it constructs the current CED $D_t$ from $y_t$ and previously issued raw forecasts. The updated components use $D_t$ and its recent trajectory to generate the CET and TRL corrections, which are added to the backbone prediction to produce $\hat Y_t^{\mathrm{EFC}}$.

EFC distinguishes immediately available error feedback from supervision that requires a complete forecasting horizon. CED can be constructed as soon as $y_t$ arrives, whereas the complete error $E_i$ becomes available only when $i+H\le t$. Only these fully matured historical origins are used for CET estimation, TRL training, and reliability evaluation. Once the error $E_i$ fully matures, TRL constructs its target using the CET correction saved at origin $i$, rather than recomputing the correction with a later estimator.

Both reliability estimates are computed using previously saved estimators and validation samples that follow their fitting or training periods. Specifically, the transport estimator $W_v$ and the TRL block $\theta_v$ used for validation satisfy:

\begin{equation}
    \operatorname{fit}_{\max}(W_v)
<
\min \mathcal V_t^\alpha,
\quad
\operatorname{train}_{\max}(\theta_v)
<
\min \mathcal V_t^\beta.
\label{eq:3}
\end{equation}

At each reliability update, validation errors must be fully matured. The CET and TRL estimators used for validation are fitted or trained on earlier, nonoverlapping forecast origins, as required by (\ref{eq:3}). All reliability computations use only information available at the current forecast time. When no eligible historical estimator or validation evidence is available, the corresponding correction remains inactive. 

\section{Experiments}
To comprehensively verify the performance and effectiveness of the proposed EFC, we conduct extensive experiments on different tasks.

\subsection{Experimental Setup}
\subsubsection{Benchmarks}
We evaluate EFC in two forecasting settings: conventional time series forecasting (TSF) \cite{TimesNet2023} and time series foundation model forecasting (TSFM) \cite{qiao2026it}. For TSF, we use ETT (ETTh1, ETTh2, ETTm1, ETTm2), Electricity, Exchange, Solar-Energy, Weather, and Traffic datasets. The details of these datasets are listed in Table \ref{tab:datasets1}. For TSFM evaluation, we adopt TIME \cite{qiao2026it}, a task-centric benchmark comprising 50 datasets and 98 forecasting tasks across diverse domains and temporal resolutions. We evaluate EFC on 92 of the 98 TIME tasks from 50 datasets. For both benchmarks, we follow their prescribed forecasting configurations and evaluation windows.

\begin{table*}[htbp]
\centering

\caption{
Descriptions of conventional time series forecasting benchmarks.
Length denotes the total number of timestamps, Variates denotes the number
of observed variables, and Split represents the number of samples in
(train, validation, test).
}

\label{tab:datasets1}

\scriptsize
\setlength{\tabcolsep}{4pt}
\renewcommand{\arraystretch}{1.2}
\begin{tabular}{>{\raggedright\arraybackslash}p{2cm}lcccccc}
\toprule
\textbf{Dataset} & \textbf{Domain} & \textbf{Length} & \textbf{Variates} & \textbf{Forecast Horizon} & \textbf{Dataset Size} & \textbf{Interval} \\
\midrule
ETTh1 & Temperature & 14,400 & 7 & \{96,192,336,720\} & (8545,2881,2881) & 1 hour\\
ETTh2 & Temperature & 14,400 & 7 & \{96,192,336,720\} & (8545,2881,2881) & 1 hour\\
ETTm1 & Temperature & 57,600 & 7 & \{96,192,336,720\} & (34465,11521,11521) & 15 min\\
ETTm2 & Temperature & 57,600 & 7 & \{96,192,336,720\} & (34465,11521,11521) & 15 min\\
Electricity & Electricity & 26,304 & 321 & \{96,192,336,720\} & (18317,2633,5261) & 1 hour\\
Exchange & Exchange Rate & 7,588 & 8 & \{96,192,336,720\} & (5120,665,1422) & 1 day\\
Traffic & Road Occupancy & 17,544 & 862 & \{96,192,336,720\} & (12185,1757,3509) & 1 hour\\
Weather & Weather & 52,696 & 21 & \{96,192,336,720\} & (36792,5271,10540) & 10 min\\
Solar-Energy & Energy & 52,179 & 137 & \{96,192,336,720\} & (36601,5161,10417) & 10 min\\
\bottomrule
\end{tabular}
\end{table*}

\subsubsection{Baselines}
We evaluate EFC on 5 state-of-the-art forecasting backbones with diverse architectures, including Transformer-based models: PatchTST \cite{PatchTST2023}, iTransformer \cite{iTransformer2023}, TimeBridge \cite{liu2025t}; MLP-based models: DLinear \cite{Ailing2023}, PatchMLP \cite{tang2024}; and 5 foundation models: VisionTS \cite{chen2025visionts}, Sundial \cite{liu2026sundial}, Chronos-Bolt \cite{ansari2024chronos}, Moirai \cite{MOIARI2024}, TimesFM-2.0 \cite{das2024decoder}.

\subsubsection{Evaluation Metrics}
For TSF benchmark, we use mean squared error (MSE) and mean absolute error (MAE). For Time benchmark, we use mean absolute scaled error (MASE) as metric, for aggregate results, we normalize task-level scores by the corresponding Seasonal Naive scores and report their geometric mean across the evaluated tasks.

\subsubsection{Implementation Details}
All experiments are implemented in PyTorch, and conducted on a single NVIDIA RTX 4090 24 GB GPU. We follow the official implementations and reported configurations of the forecasting backbones and competing methods. For conventional forecasting backbones trained in our experiments, we use Adam optimizer with an initial learning rate of $10^{-3}$ and the mean squared error (MSE) loss. Pretrained foundation models are evaluated in a zero-shot setting with their parameters frozen, only the EFC correction components are updated during deployment.

For EFC, we set the default error-history length to $R=8$, the input projection dimension to $z=32$, and the hidden dimension to $d_{hid}=32$. CET maintains a separate transport matrix for each variable, with the cumulative second-order statistics initialized using ridge regularization $S_0=\lambda I$, where $\lambda=1.0$ in default. TRL consists of an input projection, a single layer GRU, and an output projection. It is optimized using Adam optimizer with a learning rate of $10^{-3}$ and the Huber loss with $\delta=1.0$. The reliability coefficients for CET and TRL are smoothed using exponential moving averages with $\rho=0.95$.
 
CET and TRL estimate reliability using the same rolling buffer of up to \(N_H=\max(150,2H)\) fully matured forecasting origins, subsampled at two stride. Reliability recomputation begins after at least \(N_{\min}=50\) matured error pairs become available and is subsequently scheduled every 30 forecast origins. TRL performs one gradient step at an eligible checkpoint, using at most 256 evenly spaced pairs from the subsampled history. The reliability denominator constant is set to \(\epsilon=10^{-12}\). At most $\lceil N_H/30 \rceil + 8$ historical CET are retained.


\definecolor{GroupGray}{RGB}{244,245,247}
\definecolor{EFCBlue}{RGB}{239,246,251}
\definecolor{EFCText}{RGB}{35,67,91}

\newcolumntype{E}{>{\columncolor{EFCBlue}}c}
\begin{table*}[!t]
\centering
\caption{Average forecasting performance across four prediction horizons ($H\in\{96,192,336,720\}$), with a fixed look-back length of $L=96$. Lower MSE and MAE values within each Base--EFC pair are highlighted in bold.}
\label{tab:tsf_avg}
\footnotesize
\setlength{\tabcolsep}{2.4pt}
\renewcommand{\arraystretch}{1.2}
\begin{adjustbox}{width=0.95\textwidth}
\begin{tabular}{
@{}l@{\hspace{5pt}}ccEE@{\hspace{6pt}}ccEE@{\hspace{6pt}}ccEE@{\hspace{6pt}}ccEE@{\hspace{6pt}}ccEE@{}}
\toprule
\multirow{3}{*}{\textbf{Dataset}} & \multicolumn{4}{>{\columncolor{GroupGray}}c} {\textbf{PatchTST}} & \multicolumn{4}{>{\columncolor{GroupGray}}c} {\textbf{iTransformer}} & \multicolumn{4}{>{\columncolor{GroupGray}}c} {\textbf{DLinear}} & \multicolumn{4}{>{\columncolor{GroupGray}}c} {\textbf{PatchMLP}} & \multicolumn{4}{>{\columncolor{GroupGray}}c} {\textbf{TimeBridge}} \\
\cmidrule(lr){2-5} \cmidrule(lr){6-9} \cmidrule(lr){10-13} \cmidrule(lr){14-17} \cmidrule(lr){18-21}
& \multicolumn{2}{c}{Base} & \multicolumn{2}{>{\columncolor{EFCBlue}}c} {{\textbf{+EFC}}} & \multicolumn{2}{c}{Base} & \multicolumn{2}{>{\columncolor{EFCBlue}}c} {{\textbf{+EFC}}} & \multicolumn{2}{c}{Base} & \multicolumn{2}{>{\columncolor{EFCBlue}}c} {{\textbf{+EFC}}} & \multicolumn{2}{c}{Base} & \multicolumn{2}{>{\columncolor{EFCBlue}}c} {{\textbf{+EFC}}} & \multicolumn{2}{c}{Base} & \multicolumn{2}{>{\columncolor{EFCBlue}}c} {{\textbf{+EFC}}} \\
& MSE & MAE & MSE & MAE & MSE & MAE & MSE & MAE & MSE & MAE & MSE & MAE & MSE & MAE & MSE & MAE & MSE & MAE & MSE & MAE \\
\midrule
ETTh1 
& 0.486 & 0.459 & \textbf{ 0.476 } & \textbf{ 0.455 } & 0.478 & 0.457 & \textbf{ 0.473 } & \textbf{ 0.454 } & 0.484 & 0.467 & \textbf{ 0.472 } & \textbf{ 0.459 } & 0.480 & 0.454 & \textbf{ 0.473 } & \textbf{ 0.450 } & 0.479 & 0.454 & \textbf{ 0.469 } & \textbf{ 0.449 } \\ 
ETTh2 
& 0.430 & 0.436 & \textbf{ 0.420 } & \textbf{ 0.431 } & 0.428 & 0.437 & \textbf{ 0.421 } & \textbf{ 0.434 } & 0.593 & 0.533 & \textbf{ 0.556 } & \textbf{ 0.513 } & 0.454 & 0.453 & \textbf{ 0.445 } & \textbf{ 0.448 } & 0.425 & 0.434 & \textbf{ 0.416 } & \textbf{ 0.430 } \\ 
ETTm1 
& 0.390 & 0.403 & \textbf{ 0.379 } & \textbf{ 0.397 } & 0.411 & 0.412 & \textbf{ 0.399 } & \textbf{ 0.406 } & 0.407 & 0.409 & \textbf{ 0.377 } & \textbf{ 0.391 } & 0.391 & 0.398 & \textbf{ 0.380 } & \textbf{ 0.392 } & 0.393 & 0.401 & \textbf{ 0.383 } & \textbf{ 0.397 } \\ 
ETTm2 
& 0.291 & 0.334 & \textbf{ 0.284 } & \textbf{ 0.330 } & 0.302 & 0.342 & \textbf{ 0.293 } & \textbf{ 0.339 } & 0.359 & 0.404 & \textbf{ 0.330 } & \textbf{ 0.367 } & 0.295 & 0.334 & \textbf{ 0.288 } & \textbf{ 0.330 } & 0.290 & \textbf{ 0.331 } & \textbf{ 0.283 } & 0.332 \\ 
Electricity & 
0.231 & 0.310 & \textbf{ 0.217 } & \textbf{ 0.299 } & 0.192 & 0.275 & \textbf{ 0.178 } & \textbf{ 0.271 } & 0.244 & 0.331 & \textbf{ 0.219 } & \textbf{ 0.310 } & 0.217 & 0.303 & \textbf{ 0.208 } & \textbf{ 0.296 } & 0.183 & 0.272 & \textbf{ 0.179 } & \textbf{ 0.268 } \\ 
Exchange 
& 0.320 & 0.401 & \textbf{ 0.317 } & \textbf{ 0.397 } & 0.321 & 0.404 & \textbf{ 0.321 } & \textbf{ 0.401 } & 0.317 & 0.405 & \textbf{ 0.312 } & \textbf{ 0.399 } & 0.299 & 0.393 & \textbf{ 0.295 } & \textbf{ 0.392 } & 0.413 & 0.463 & \textbf{ 0.408 } & \textbf{ 0.459 } \\ 
Solar 
& 0.262 & 0.301 & \textbf{ 0.235 } & \textbf{ 0.285 } & 0.230 & 0.260 & \textbf{ 0.226 } & \textbf{ 0.254 } & 0.322 & 0.398 & \textbf{ 0.266 } & \textbf{ 0.329 } & 0.288 & 0.298 & \textbf{ 0.244 } & \textbf{ 0.283 } & 0.232 & \textbf{ 0.269 } & \textbf{ 0.231 } & 0.271 \\ 
Traffic 
& 0.522 & 0.374 & \textbf{ 0.509 } & \textbf{ 0.369 } & 0.408 & 0.300 & \textbf{ 0.403 } & \textbf{ 0.295 } & 0.657 & 0.439 & \textbf{ 0.597 } & \textbf{ 0.416 } & 0.516 & 0.384 & \textbf{ 0.505 } & \textbf{ 0.379 } & 0.371 & 0.287 & \textbf{ 0.367 } & \textbf{ 0.284 } \\ 
Weather 
& 0.258 & 0.278 & \textbf{ 0.245 } & \textbf{ 0.270 } & 0.260 & 0.278 & \textbf{ 0.249 } & \textbf{ 0.271 } & 0.266 & 0.315 & \textbf{ 0.232 } & \textbf{ 0.276 } & 0.251 & 0.275 & \textbf{ 0.241 } & \textbf{ 0.268 } & 0.258 & 0.279 & \textbf{ 0.245 } & \textbf{ 0.271 } \\
\bottomrule
\end{tabular}
\end{adjustbox}
\end{table*}

\subsection{Main Results}

We present the main results on 9 real-world datasets for five SOTA forecasting models in Table \ref{tab:tsf_avg}, where the lower MSE and MAE indicate better forecasting performance. EFC reduces the average MSE and MAE across most cases. Specifically, EFC achieves a 4.16\% average reduction in MSE. The improvement is broad across different forecasting architectures, with the largest average reduction achieved on DLinear with 8.37\%. Its gains are particularly notable on complex datasets such as Solar (17.39\%) and Traffic (9.13\%). Overall, these results demonstrate that error feedback can improve the forecasts of diverse frozen backbones, with the magnitude of improvement depending on the model and dataset. 

\begin{table}[htbp]
\centering
\caption{
Average forecasting performance on time series foundation models
under the TIME benchmark. Lower normalized MASE indicates better forecasting accuracy. Boldface denotes the better result.
}
\label{tab:tsfm_results}
\scriptsize
\setlength{\tabcolsep}{5pt}
\renewcommand{\arraystretch}{1.15}
\begin{tabular}{
>{\raggedright\arraybackslash}p{1.7cm}lcccc}
\toprule
\rowcolor{GroupGray}
{\textbf{Model}} & \multicolumn{2}{c}{\textbf{Normalized MASE $\downarrow$}} & {\textbf{Reduction}} & {\textbf{Task Win}} \\
\cmidrule(lr){2-3}
 & Frozen & \cellcolor{EFCBlue}\textbf{+EFC} & (\%) & /92 \\
\midrule
VisionTS&0.808&\cellcolor{EFCBlue}\textbf{0.776}&4.0&80/92\\
Sundial&0.746&\cellcolor{EFCBlue}\textbf{0.731}&1.9&78/92\\
Chronos-Bolt&0.728&\cellcolor{EFCBlue}\textbf{0.701}&3.7&71/92\\
Moirai&0.830&\cellcolor{EFCBlue}\textbf{0.791}&4.7&70/92\\
TimesFM-2.0&0.713&\cellcolor{EFCBlue}\textbf{0.652}&8.5&68/92\\
\midrule
Average&0.765&\cellcolor{EFCBlue}\textbf{0.730}&4.5&73.4/92\\
\bottomrule
\end{tabular}
\end{table}

Meanwhile, we evaluate EFC with pretrained foundation models under the TIME benchmark for zero-shot forecasting. As listed in Table \ref{tab:tsfm_results}, EFC improves the forecasting accuracy of different TSFMs by using only deployment-time revealed errors, reducing the normalized MASE from 0.765 to 0.730 on average. The task-level results further illustrate the breadth of these improvements. Across the five foundation models, EFC improves 68--80 of the 92 evaluated TIME tasks, indicating that the aggregate gains are not limited to a few favorable cases. However, 12--24 tasks per model show no improvement, suggesting that the effectiveness of error feedback varies across forecasting tasks. One possible explanation is that revealed historical errors do not always provide sufficiently reliable information about future residuals. These results demonstrate improvements across a broad range of tasks, although EFC does not improve every evaluated task.

\subsection{Comparison with other Methods}
To further validate the effectiveness of EFC, we compare it with PIR \cite{PIR2025}, COSA \cite{im2026cosa}, PETSA \cite{medeiros2025}, and TAFAS \cite{kim2025} on four datasets using iTransformer as backbone. As shown in Table \ref{tab:tta_comparison}, EFC achieves superior forecasting performance in most settings. Its advantages are particularly evident at short and medium prediction horizons on Electricity and ETTh1, where improvements are observed for both metrics. By using previously revealed forecasting errors as feedback, EFC provides an effective way to correct future predictions without updating the forecasting backbone, while remaining competitive with existing adaptation methods.

\definecolor{HeaderGray}{RGB}{244,245,247}
\definecolor{EFCBlue}{RGB}{239,246,251}

\begin{table*}[!t]
\centering
\caption{
Comparison with representative post-hoc forecast revision and test-time adaptation methods on Electricity, ETTh1, Weather, Traffic datasets. We use iTransformer as the backbone. The best results are highlighted in bold, and the second-best results are underlined.
}
\label{tab:tta_comparison}
\scriptsize
\setlength{\tabcolsep}{2.8pt}
\setlength{\extrarowheight}{0.65pt}
\renewcommand{\arraystretch}{1.15}
\begin{tabular}{@{}l c *{16}{c}@{}}
\toprule
\rowcolor{HeaderGray}
\textbf{Method} & \textbf{Metric}
& \multicolumn{4}{c}{\textbf{Electricity}}
& \multicolumn{4}{c}{\textbf{ETTh1}}
& \multicolumn{4}{c}{\textbf{Weather}}
& \multicolumn{4}{c}{\textbf{Traffic}} \\
\arrayrulecolor{HeaderGray}%
\specialrule
{\dimexpr\aboverulesep+\cmidrulewidth+\belowrulesep+0.5pt\relax}
{0pt}
{-\dimexpr\aboverulesep+\cmidrulewidth+\belowrulesep+0.5pt\relax}%
\arrayrulecolor{black}%
\cmidrule(lr){3-6} \cmidrule(lr){7-10} \cmidrule(lr){11-14} \cmidrule(lr){15-18}
\rowcolor{HeaderGray}
& & 96 & 192 & 336 & 720 & 96 & 192 & 336 & 720 & 96 & 192 & 336 & 720 & 96 & 192 & 336 & 720 \\
\midrule
\multirow{2}{*}{Backbone} & MSE
& 0.152 & \underline{0.166} & 0.181 & 0.268
& \underline{0.392} & \underline{0.451} & \underline{0.516} & \textbf{0.552}
& 0.176 & 0.225 & 0.281 & 0.359
& \underline{0.396} & 0.420 & \underline{0.392} & \underline{0.422} \\
& MAE
& \underline{0.243} & \underline{0.256} & \underline{0.273} & 0.330
& 0.407 & \underline{0.442} & \underline{0.471} & \textbf{0.507}
& 0.214 & 0.256 & \underline{0.296} & 0.349
& 0.271 & 0.281 & 0.316 & 0.333 \\
\midrule
\multirow{2}{*}{PIR} & MSE
& \underline{0.152} & 0.169 & \underline{0.181} & 0.223
& 0.396 & 0.464 & 0.522 & \underline{0.601}
& 0.171 & 0.218 & 0.279 & 0.362
& \textbf{0.389} & \textbf{0.414} & 0.427 & 0.455 \\
& MAE
& 0.245 & 0.261 & 0.277 & \textbf{0.312}
& \underline{0.402} & 0.456 & 0.501 & 0.597
& 0.212 & 0.253 & 0.300 & 0.356
& \textbf{0.267} & \underline{0.278} & 0.323 & 0.336 \\
\addlinespace[2pt]
\multirow{2}{*}{COSA} & MSE
& 0.162 & 0.175 & 0.188 & \textbf{0.212}
& 0.439 & 0.499 & 0.557 & 0.639
& 0.168 & 0.215 & 0.272 & \textbf{0.342}
& 0.415 & 0.439 & 0.454 & 0.482 \\
& MAE
& 0.253 & 0.264 & 0.279 & 0.325
& 0.450 & 0.494 & 0.531 & \underline{0.587}
& 0.211 & 0.254 & 0.299 & \textbf{0.345}
& 0.286 & 0.295 & \textbf{0.301} & \textbf{0.317} \\
\addlinespace[2pt]
\multirow{2}{*}{PETSA} & MSE
& 0.160 & 0.174 & 0.190 & 0.231
& 0.436 & 0.496 & 0.567 & 0.660
& 0.167 & 0.211 & \underline{0.265} & \underline{0.347}
& 0.414 & 0.438 & 0.453 & 0.485 \\
& MAE
& 0.253 & 0.266 & 0.284 & 0.331
& 0.450 & 0.493 & 0.538 & 0.602
& \underline{0.210} & \underline{0.253} & 0.297 & 0.354
& 0.286 & 0.295 & 0.317 & 0.319 \\
\addlinespace[2pt]
\multirow{2}{*}{TAFAS} & MSE
& 0.159 & 0.173 & 0.190 & 0.230
& 0.439 & 0.495 & 0.572 & 0.666
& \underline{0.166} & \underline{0.210} & \textbf{0.262} & 0.349
& 0.414 & 0.437 & 0.454 & 0.493 \\
& MAE
& 0.252 & 0.265 & 0.283 & 0.335
& 0.450 & 0.493 & 0.545 & 0.607
& 0.212 & 0.255 & 0.297 & 0.360
& 0.286 & 0.294 & \textbf{0.301} & 0.326 \\
\midrule
\rowcolor{EFCBlue}
 & MSE
& \textbf{0.150} & \textbf{0.163} & \textbf{0.179} & \underline{0.217}
& \textbf{0.389} & \textbf{0.444} & \textbf{0.509} & \textbf{0.552}
& \textbf{0.165} & \textbf{0.209} & 0.266 & 0.356
& \textbf{0.389} & \underline{0.415} & \textbf{0.388} & \textbf{0.420} \\
\rowcolor{EFCBlue}
\multirow{-2}{*}{\textbf{EFC (Ours)}}& MAE
& \textbf{0.242} & \textbf{0.255} & \textbf{0.272} & \underline{0.317}
& \textbf{0.401} & \textbf{0.440} & \textbf{0.469} & \textbf{0.507}
& \textbf{0.205} & \textbf{0.248} & \textbf{0.285} & \underline{0.346}
& \underline{0.269} & \textbf{0.277} & \underline{0.315} & \underline{0.318} \\
\bottomrule
\end{tabular}
\end{table*}

\subsection{Ablation Studies}
To validate the contribution of each component in EFC, we conduct ablation studies using DLinear on the Weather dataset. We evaluate three aspects of EFC: the error representation introduced by CED, the reliable error transport mechanism in CET, and the residual prediction capability of TRL. The results are reported in Table \ref{tab:ced_ablation}--\ref{tab:trl_ablation}.

For Causal Error Diagonal (CED), we compare different representations: \ding{172} without error feedback. \ding{173} the latest fully mature error $E_{t-H}$, which provides complete but delayed feedback. \ding{174} the one-step revealed error $e_{t-1,1}$, which provides immediate feedback but lacks multi-lead information. And \ding{175} the complete CED representation. As shown in Table \ref{tab:ced_ablation}, both matured errors and one-step errors improve over the frozen backbone, indicating that previously revealed forecasting errors contain useful information for future correction. However, their improvements are limited compared with the full CED representation. The matured error suffers from the inherent $H$-step delay, while the one-step error only captures a single lead time relationship. CED achieves the best performance, reducing MSE in an 18.46\% reduction. This demonstrates that effective error feedback benefits from simultaneously preserving temporal freshness and cross-origin multi-lead error structures.


\begin{table}[htbp]
\centering
\caption{Ablation study of Causal Error Diagonal (CED).}
\label{tab:ced_ablation}
\scriptsize
\setlength{\tabcolsep}{5pt}
\renewcommand{\arraystretch}{1.15}
\resizebox{\columnwidth}{!}{
\begin{tabular}{
>{\raggedright\arraybackslash}p{2.5cm}cc|cc}
\toprule
\rowcolor{HeaderGray} 
\makecell{\textbf{Error}\\\textbf{Representation}} & \textbf{Delay} & \textbf{Multi-lead} & \textbf{MSE $\downarrow$} & \textbf{$\Delta$MSE $\downarrow$} \\
\cmidrule(lr){1-3} \cmidrule(l){4-5}
\ding{172} w/o CED & -- & $\times$ & 0.195 & -- \\
\ding{173} Matured error $E_{t-H}$ & $H$ & $\checkmark$ & 0.164 & -15.90\% \\
\ding{174} One-step error & 0 & $\times$ & 0.171 & -12.31\%
\\
\rowcolor{EFCBlue}
{\textbf{\ding{175} CED}} & \textbf{0} & \textbf{$\checkmark$} & \textbf{0.159} & \textbf{-18.46\%} \\
\bottomrule
\end{tabular}
}
\end{table}

For Cumulative Error Transport (CET), we evaluate: \ding{172} directly applies CED as the correction signal without learning an error transport relation. \ding{173} uses the cumulative transport matrix while removing online reliability control. \ding{174} adds the temporally out-of-sample reliability estimation. And \ding{175} uses the complete CET design with both reliability estimation and EMA smoothing. As listed in Table \ref{tab:cet_ablation}, directly reusing CED as a correction signal substantially degrades performance, increasing MSE by 62.56\%. This indicates that revealed errors describe the forecaster's recent failure state but cannot be directly interpreted as future corrections. Learning the transport relation significantly improves the results, reducing MSE by 14.87\%, which verifies the necessity of estimating how past errors transfer to future residuals. Further out-of-sample reliability estimation provides additional improvements, with the complete CET achieving the best performance. These results demonstrate that reliable error feedback requires both error transport and online correction control.

\begin{table}[htbp]
\centering
\caption{Ablation study of Cumulative Error Transport (CET).}
\label{tab:cet_ablation}
\scriptsize
\setlength{\tabcolsep}{4.5pt}
\renewcommand{\arraystretch}{1.15}
\resizebox{\columnwidth}{!}{
\begin{tabular}{
>{\raggedright\arraybackslash}p{2.1cm}ccc|cc}
\toprule
\rowcolor{HeaderGray} 
\textbf{Variant} & \textbf{Transport} & \textbf{OOS $\alpha$} & \textbf{EMA} & \textbf{MSE $\downarrow$} & \textbf{$\Delta$MSE $\downarrow$} \\
\cmidrule(lr){1-4} \cmidrule(l){5-6}
\ding{172} Direct CED & $\times$ & -- & -- & 0.317 & 62.56\% \\
\ding{173} Raw Transport & $\checkmark$ & $\times$ & $\times$ & 0.166 & -14.87\% \\
\ding{174} + OOS Reliability & $\checkmark$ & $\checkmark$ & $\times$ & 0.161 & -17.44\% \\
\rowcolor{EFCBlue}
{\ding{175} \textbf{Full CET}} & \textbf{$\checkmark$} & \textbf{$\checkmark$} & \textbf{$\checkmark$} & \textbf{0.159} & \textbf{-18.46\%} \\
\bottomrule
\end{tabular}
}
\end{table}

For Temporal Residual Learning (TRL), we evaluate: \ding{172} only CET without TRL, \ding{173} removes temporal history by setting $R=1$. \ding{174} replaces the residual target with the full forecasting error. \ding{175} removes the reliability coefficient by fixing $\beta=1$. And \ding{176} the complete TRL design. As shown in Table \ref{tab:trl_ablation}, the complete TRL design reduces MSE from 0.170 for CET only to 0.159, an additional relative reduction of 6.47\%. Removing temporal context decreases the improvement, indicating that residual errors contain temporal dependencies beyond the latest error state. Predicting the full error also weakens the performance, suggesting that residual learning is more effective when focusing on the remaining correction after CET rather than relearning already captured error patterns. Removing OOS reliability slightly reduces the gain, confirming the importance of controlling residual corrections during deployment. Overall, these results verify that temporally structured residual information remains available after cumulative error transport.

\begin{table}[htbp]
\centering
\caption{Ablation study of Temporal Residual Learning (TRL).}
\label{tab:trl_ablation}
\scriptsize
\setlength{\tabcolsep}{3.5pt}
\renewcommand{\arraystretch}{1.15}
\resizebox{\columnwidth}{!}{
\begin{tabular}{
>{\raggedright\arraybackslash}p{1.9cm}ccc|cc}
\toprule
\rowcolor{HeaderGray}
\textbf{Variant} & \makecell{\textbf{Temporal}\\\textbf{Context}} & \makecell{\textbf{Residual}\\\textbf{Target}} & \textbf{OOS $\beta$} & \textbf{MSE $\downarrow$} & \textbf{$\Delta$over CET $\downarrow$} \\
\cmidrule(lr){1-4} \cmidrule(l){5-6}
\ding{172} CET only & -- & -- & -- & 0.170 & -- \\
\ding{173} $R=1$ & $\times$ & $\checkmark$ & $\checkmark$ & 0.162 & -4.71\% \\
\ding{174} Full-error target & $\checkmark$ & $\times$ & $\checkmark$ & 0.166 & -2.35\% \\
\ding{175} $\beta=1$ & $\checkmark$ & $\checkmark$ & $\times$ & 0.160 & -5.88\% \\
\rowcolor{EFCBlue}
{\ding{176} \textbf{Full TRL}} & \textbf{$\checkmark$} & \textbf{$\checkmark$} & \textbf{$\checkmark$} & \textbf{0.159} & \textbf{-6.47\%} \\
\bottomrule
\end{tabular}
}
\end{table}

\subsection{Method Analysis}
\subsubsection{Error Feedback Signal}
\begin{figure}[htbp]
    \centering
    \includegraphics[width=0.9\columnwidth]{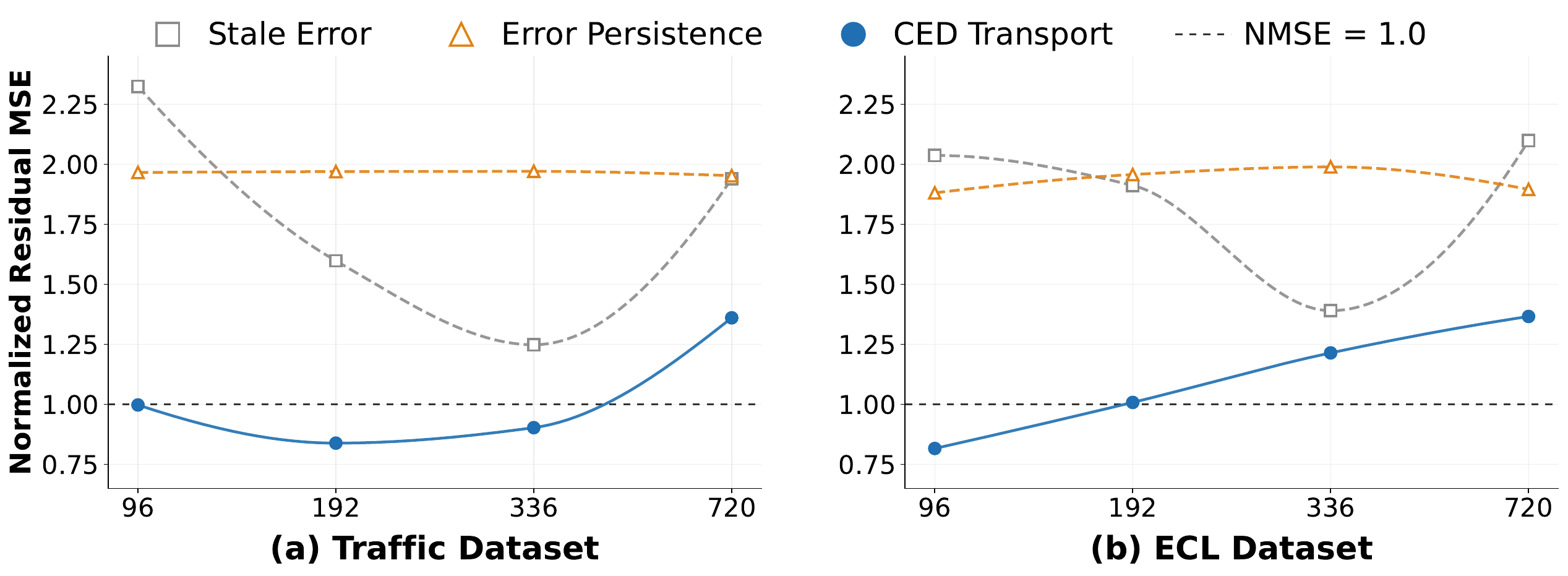}
    \caption{Method analysis of error feedback signal. We use DLinear as backbone, with results on Traffic in (a) and ECL in (b). Normalized residual MSE is computed relative to zero correction, with a reference value of 1.0. Lower values indicate better prediction of future forecasting errors.}
    \label{fig:efs}
\end{figure}

To investigate whether revealed forecasting errors contain predictive information for future forecast deviations, we evaluate different error feedback signals using normalized residual MSE. Let $E_t$ denote the frozen backbone error and $\Delta_t$ a candidate error correction. For the set of evaluated forecast origins $\mathcal{I}$, we define the residual MSE as follows:
\begin{equation}
    \mathcal{M}(\Delta)=
\frac{1}{|\mathcal{I}|HC}
\sum{t\in\mathcal{I}}
|E_t-\Delta_t|_F^2.
\end{equation}
We compare stale matured errors, direct error persistence, and the proposed CED transport on DLinear with Electricity and Traffic datasets. The fully matured error provides a complete residual vector, but its availability is delayed by $H$ steps. Direct persistence uses more recent feedback, yet assumes that the revealed error can be carried forward without accounting for changes across forecast lead times. In contrast, CED collects errors from overlapping historical origins that are verified by the same observation, and transport estimates how this cross-origin state relates to subsequent residuals. As shown in Fig. \ref{fig:efs}, CED transport achieves lower residual errors than the two alternatives in most forecasting horizons, with normalized MSE remaining below 1 for short and medium horizons. In contrast, matured errors suffer from delayed availability, while directly reusing revealed errors fails to capture the evolving relationship between past and future forecast deviations. These results indicate that deployment time forecast errors are not merely historical evaluation signals, but contain structured information that can be transformed into useful feedback for future correction.

\subsubsection{Cumulative Transport}

To analyze the stability of error transport estimation during deployment, we compare different online estimators while keeping the same CED representation: Online SGD, which updates the transport relation through single-step gradient optimization; Sliding RLS, which estimates the relation using a fixed-size recent error window; Forgetting RLS, which gradually discounts older observations through a forgetting factor; and Cumulative RLS, which continuously incorporates all matured error observations. These estimators are evaluated based on correction gain, win rate, and numerical stability. We report correction gain relative MSE change compared with the frozen forecaster, where negative values indicate improvement. And numerical stability indicates whether the estimator produces stable updates without divergence or excessive oscillation during deployment. As shown in Table \ref{tab:transport_stability}, Online SGD fails to provide reliable correction and introduces additional errors, with a positive MSE increase of 26.40\% and zero win rate, indicating that direct online updates are sensitive to noisy deployment feedback. Sliding RLS performs better than Online SGD but still increases MSE by 7.11\% relative to the frozen backbone and exhibits oscillating updates. Forgetting RLS achieves stable correction, while cumulative RLS further improves both effectiveness and consistency, reducing MSE by 9.72\% with an 80.9\% win rate. These results demonstrate that accumulating matured error observations provides a more stable basis for estimating the error transport relationship under evolving deployment conditions.

The estimator comparison also clarifies the role of cumulative estimation under non-stationarity. CET does not model the observed series itself; it estimates a mapping from revealed forecasting errors to future residuals. Even as the data evolve, the CED input changes with the forecaster's recent error state, while cumulative estimation pools fully matured historical pairs to reduce sensitivity to noisy updates. This is useful when the error-transport relation remains sufficiently persistent, consistent with the stability observed in our experiments. When the error-transport relationship changes abruptly, cumulative estimation may respond slowly because earlier matured samples remain in the sufficient statistics. Temporal out-of-sample reliability can reduce the strength of a correction when its historical validation evidence weakens, but does not directly accelerate the adaptation of the transport matrix. Adaptive forgetting offers a potential direction for balancing estimation stability and responsiveness under rapid changes.

\begin{table}[htbp]
\centering
\caption{
Stability analysis of different error transport estimators.
Results are evaluated across three forecasting backbones and four datasets.
Values in parentheses denote median correction gain across deployment steps.
}
\label{tab:transport_stability}
\scriptsize
\setlength{\tabcolsep}{5pt}
\renewcommand{\arraystretch}{1.15}
\begin{tabular}{
>{\raggedright\arraybackslash}p{2.5cm}ccc}
\toprule
\rowcolor{HeaderGray}
\textbf{Transport Estimator}& \makecell{\textbf{Correction Gain}\\\textbf{$\Delta$MSE $\downarrow$}} & \textbf{Win Rate}$\uparrow$ & \makecell{\textbf{Numerical}\\ \textbf{Stability}} \\
\midrule
Online SGD & +26.40 (med +14.20) & 0.000 & Unstable \\
Sliding RLS & +7.11 (med +1.66) & 0.458 & Oscillating \\
Forgetting RLS & -5.39 (med -5.15) & 0.583 & Stable \\
\rowcolor{EFCBlue}
{\textbf{Cumulative RLS(Ours)}} & \textbf{-9.72} \textbf{(med -11.43)} & \textbf{0.809} & \textbf{Stable} \\
\bottomrule
\end{tabular}
\end{table}

\subsubsection{Residual Predictability}
\begin{figure}[htbp]
    \centering
    \includegraphics[width=\columnwidth]{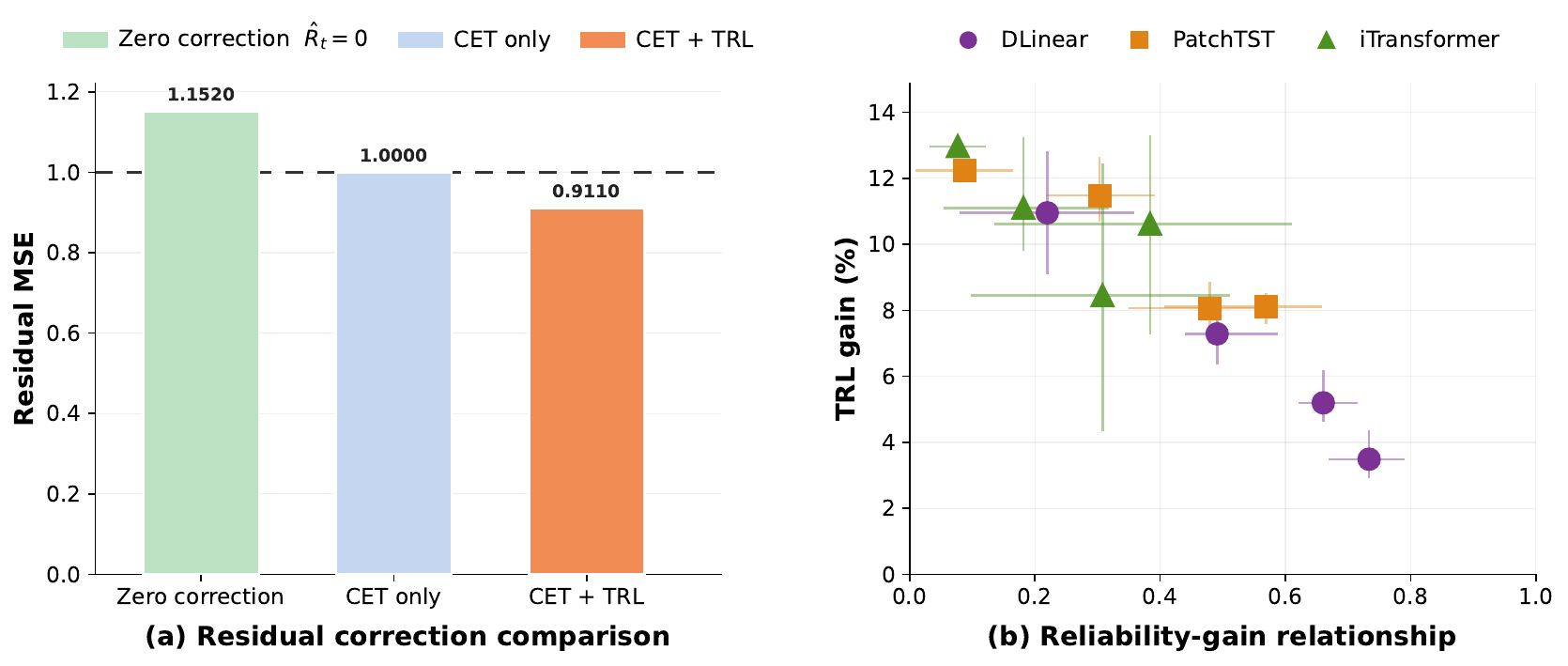}
    \caption{
    Analysis of residual predictability after CET.
    (a) Residual error reduction brought by TRL. Residual MSE normalized by CET-only correction (CET only = 1.0), comparing zero correction, CET only, and CET+TRL.
    (b) Correlation between CET reliability and TRL gains.
    }
    \label{fig:residual_predictability}
\end{figure}

To investigate whether the residual errors after CET remain predictable, we analyze the additional correction capability provided by TRL. As shown in Fig. \ref{fig:residual_predictability}a, we use CET-only correction as the normalization reference. Specifically, the normalized residual MSE is defined as:${\mathcal{M}(\Delta)}/{\mathcal{M}(\Delta^{\mathrm{CET}})}$, where $\Delta^{\mathrm{CET}}$ denotes the deployed CET correction. Under this normalization, CET only has a reference value of $1.000$, while zero correction and CET+TRL yield $1.152$ and $0.911$, respectively. The reduction from $1.000$ to $0.911$ corresponds to an additional $8.9\%$ reduction in residual MSE relative to CET alone. This result provides direct evidence that TRL can predict part of the error remaining after the first-stage CET correction. Then we further examine the relationship between CET reliability and TRL gains across different backbones and forecasting horizons. The results in Fig. \ref{fig:residual_predictability}b show a consistent tendency that lower CET reliability corresponds to larger TRL improvements. The average TRL gain increases from 3.49\% to 10.95\% for DLinear as the average reliability decreases from 0.73 to 0.22, while similar trends are observed for PatchTST and iTransformer. These observations demonstrate that TRL effectively complements CET by modeling the residual structures left unexplained by error transport.

\subsubsection{Efficiency Analysis}
We further conduct the efficiency analysis for the backbone with and without EFC. EFC keeps the forecasting backbone frozen and only introduces lightweight error correction components. Table \ref{tab:efficiency} reports the additional parameters and runtime overhead introduced by EFC across different forecasting backbones. EFC adds $0.034$--$0.051$ million parameters across the evaluated models. The relative increase depends strongly on backbone size: it reaches $45.9\%$ for DLinear, whose baseline is particularly small, but is no more than $0.2\%$ for iTransformer, TimeXer, and TimeBridge. Although the relative parameter increase is more noticeable for extremely lightweight models such as DLinear, the absolute additional parameters remain small. A similar distinction appears in runtime. The absolute increase remains below $1$ ms per iteration for all five backbones, whereas the relative increase ranges from $3.2\%$ for TimeBridge to $45.6\%$ for DLinear. Thus, EFC has a modest absolute inference cost in these settings. These results demonstrate that EFC achieves effective deployment-time correction with limited computational overhead.


\begin{table}[htbp]
\centering
\caption{
Efficiency analysis of EFC under online inference on Traffic dataset.
The overhead is computed relative to the frozen backbone.
}
\label{tab:efficiency}
\scriptsize
\setlength{\tabcolsep}{5pt}
\begin{tabular}{lcccccc}
\toprule

\multirow{2}{*}{\textbf{Backbone}} & \multicolumn{3}{c}{\textbf{Parameters}} & \multicolumn{3}{c}{\textbf{Inference Time}} \\
\cmidrule(lr){2-4} \cmidrule(lr){5-7}
& Base & \cellcolor{EFCBlue}+EFC & $\Delta$ & Base & \cellcolor{EFCBlue}+EFC & $\Delta$ \\
& (M) & \cellcolor{EFCBlue}(M) & (\%) & (ms/iter) & \cellcolor{EFCBlue}(ms/iter) & (\%) \\
\midrule
DLinear & 0.074 & \cellcolor{EFCBlue}0.108 & 45.9 & 0.630 & \cellcolor{EFCBlue}0.917 & 45.6\\
PatchTST & 2.189 & \cellcolor{EFCBlue}2.240 & 2.3 & 2.151 & \cellcolor{EFCBlue}2.529 & 17.6\\
iTransformer & 25.647 & \cellcolor{EFCBlue}25.698 & 0.2 & 1.451 & \cellcolor{EFCBlue}1.714 & 18.1\\
TimeXer & 34.933 & \cellcolor{EFCBlue}34.984 & 0.1 & 6.400 & \cellcolor{EFCBlue}7.122 & 11.3\\
TimeBridge & 77.290 & \cellcolor{EFCBlue}77.340 & 0.1 & 16.504 & \cellcolor{EFCBlue}17.028 & 3.2\\
\bottomrule
\end{tabular}
\end{table}

\begin{figure*}[!t]
    \centering
    \subfloat[Sensitivity analysis of the error history length $R$]{%
        \includegraphics[width=0.76\textwidth]{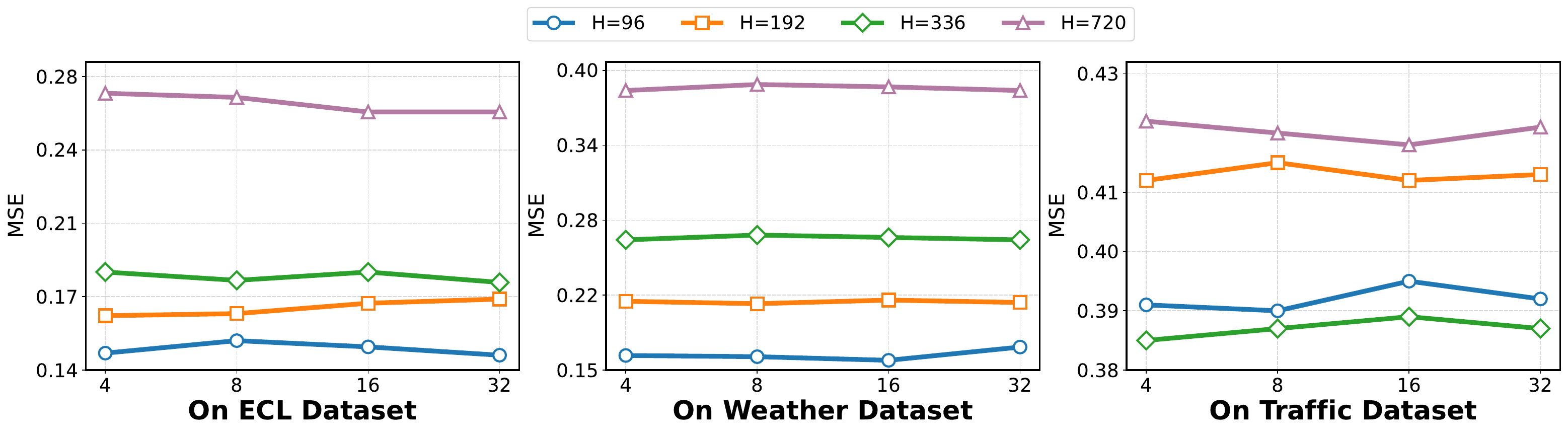}%
        \label{fig:sens_R}%
    }\\[1ex]
    \subfloat[Sensitivity analysis of the input projection dimension $z$]{%
        \includegraphics[width=0.76\textwidth]{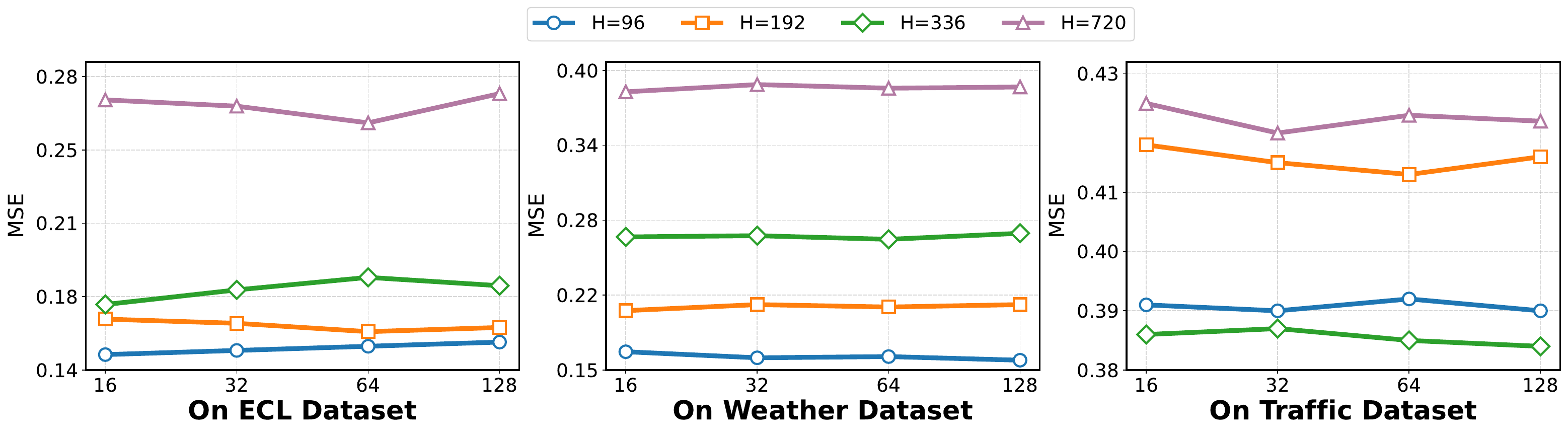}%
        \label{fig:sens_z}%
    }\\[1ex]
    \subfloat[Sensitivity analysis of the EMA smoothing coefficient $\rho$]{%
        \includegraphics[width=0.76\textwidth]{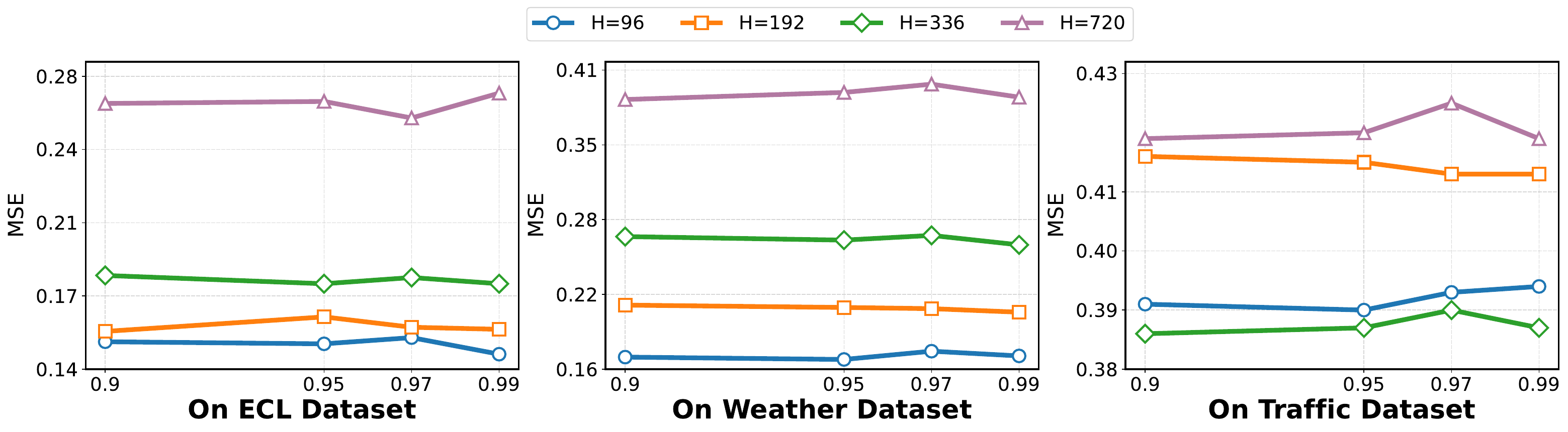}%
        \label{fig:sens_rho}%
    }\\[1ex]
    \subfloat[Sensitivity analysis of the GRU hidden dimension $d_{\mathrm{hid}}$]{%
        \includegraphics[width=0.76\textwidth]{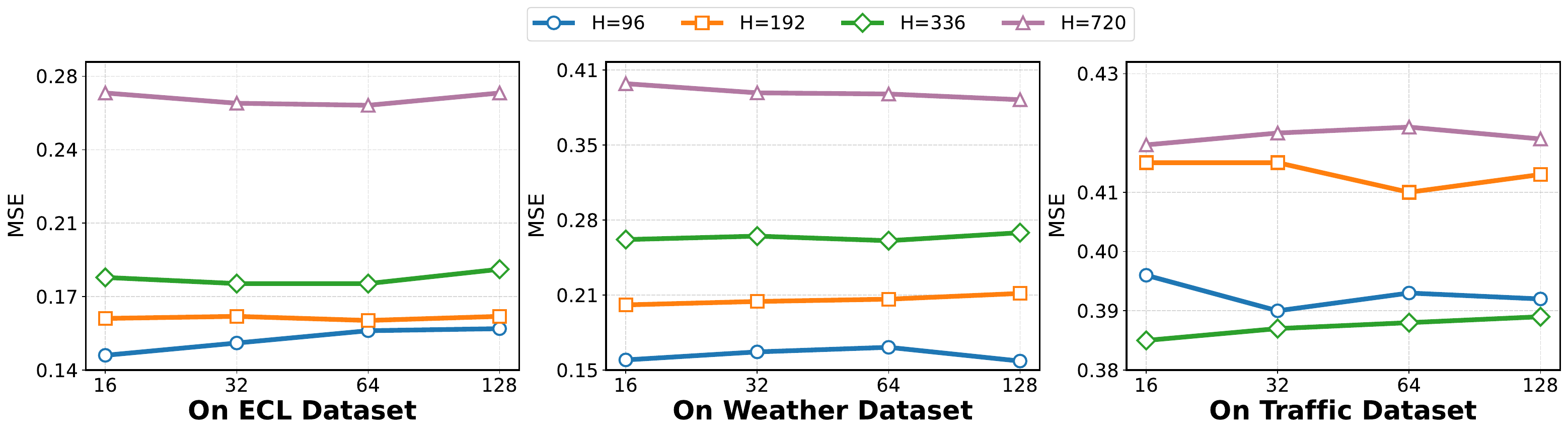}%
        \label{fig:sens_hdim}%
    }
    \caption{Hyperparameter sensitivity analysis.}
    \label{fig:sensitivity}
\end{figure*}

\begin{figure*}[!t]
    \centering
    \subfloat[Visualization on Electricity dataset]{%
        \begin{minipage}{0.31\textwidth}
            \includegraphics[width=\textwidth]{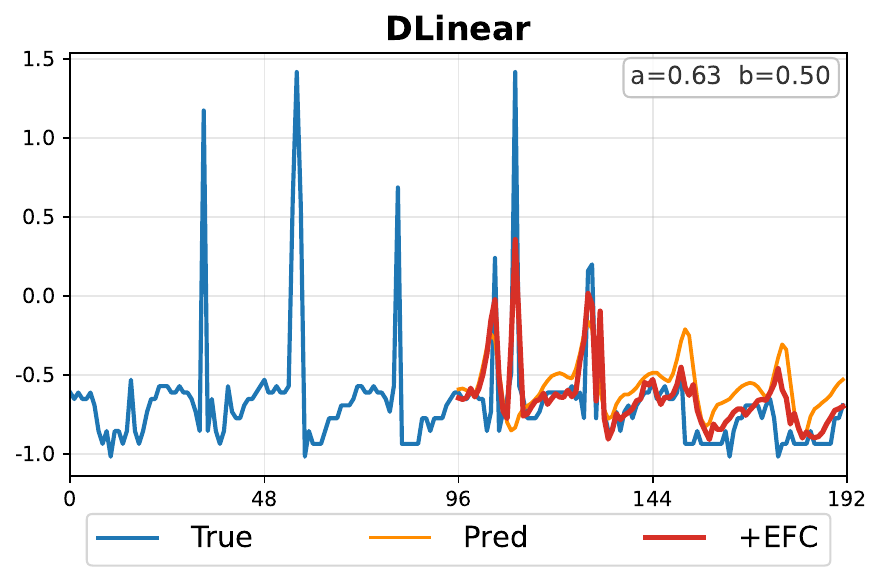}
        \end{minipage}%
        \hspace{-2pt}%
        \begin{minipage}{0.31\textwidth}
            \includegraphics[width=\textwidth]{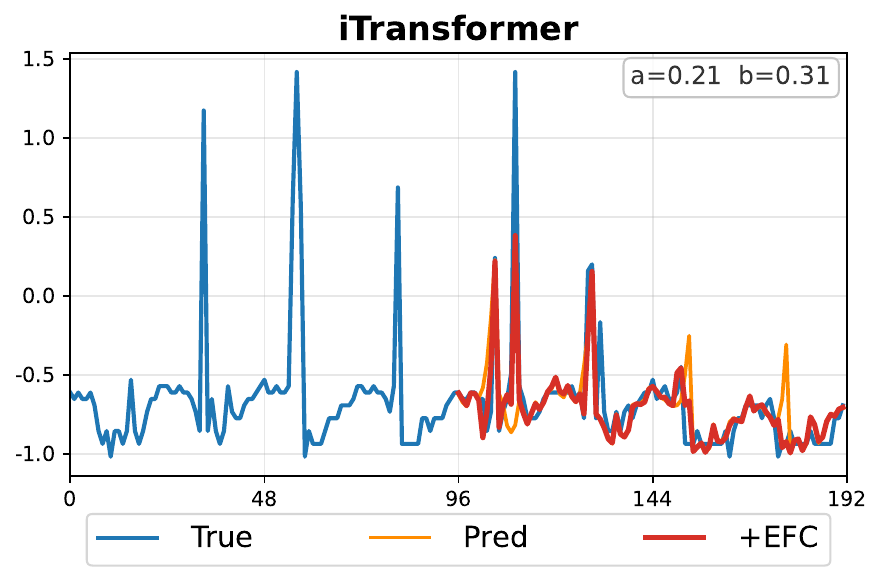}
        \end{minipage}%
        \hspace{-2pt}%
        \begin{minipage}{0.31\textwidth}
            \includegraphics[width=\textwidth]{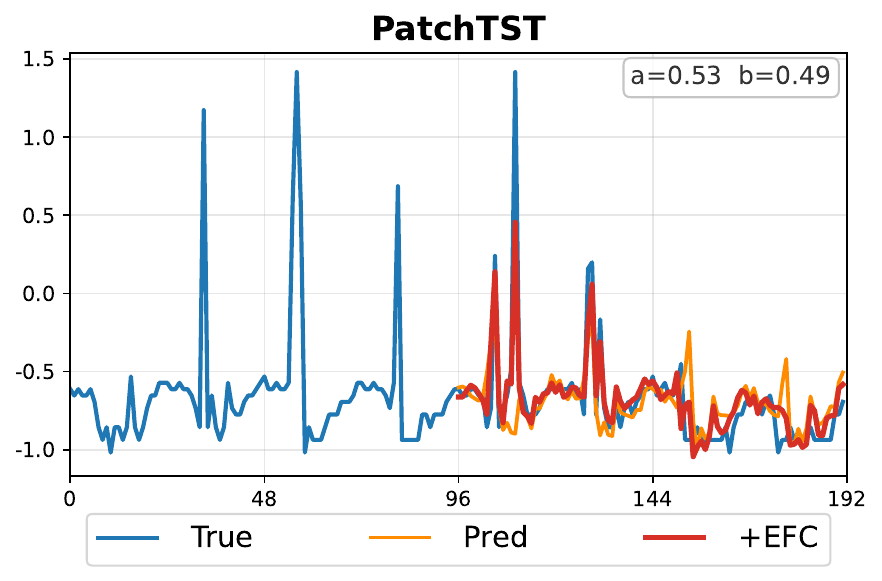}
        \end{minipage}%
        \label{fig:ecl_vis}%
    }
    \\
    \subfloat[Visualization on Traffic dataset]{%
        \begin{minipage}{0.31\textwidth}
            \includegraphics[width=\textwidth]{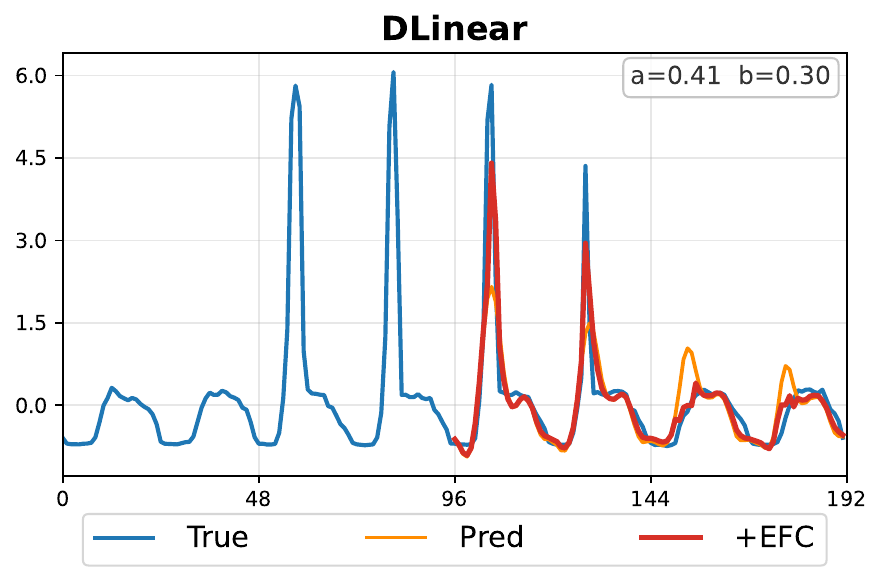}
        \end{minipage}%
        \hspace{-2pt}%
        \begin{minipage}{0.31\textwidth}
            \includegraphics[width=\textwidth]{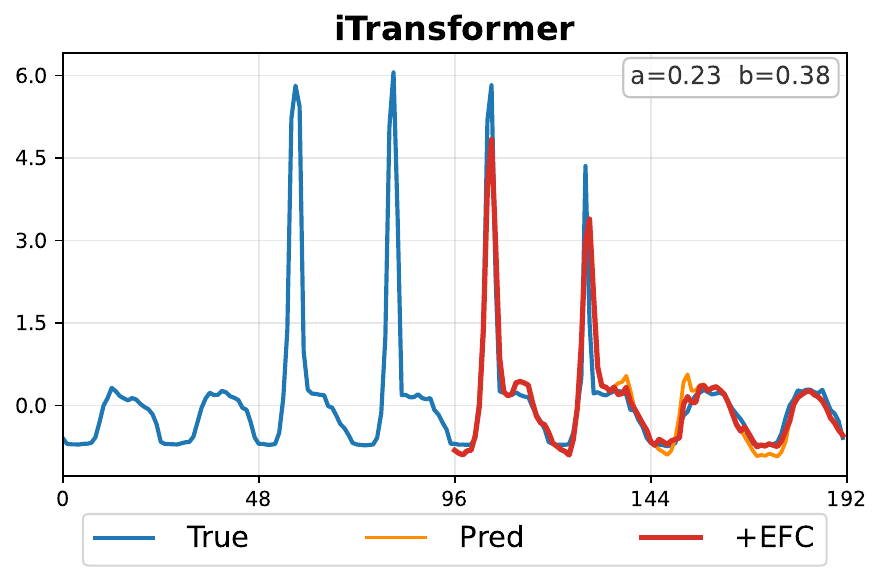}
        \end{minipage}%
        \hspace{-2pt}%
        \begin{minipage}{0.31\textwidth}
            \includegraphics[width=\textwidth]{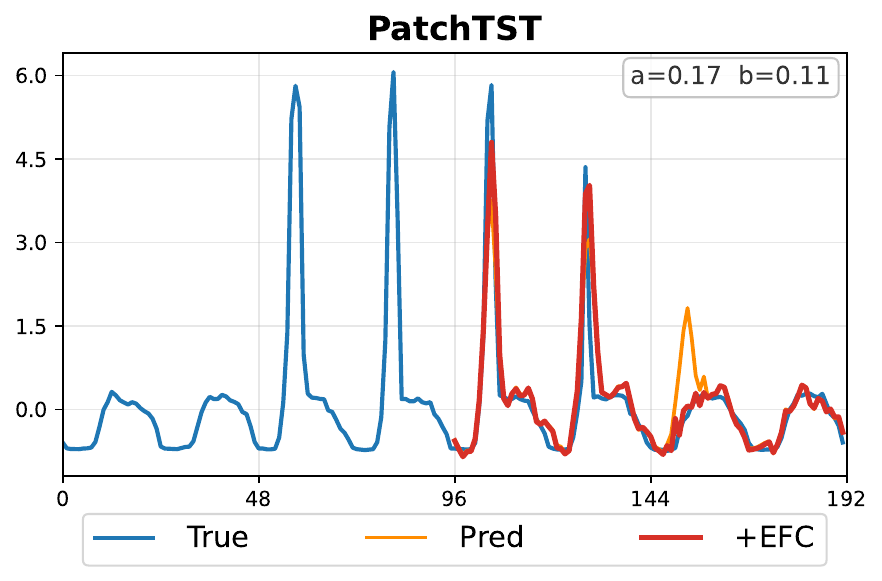}
        \end{minipage}%
        \label{fig:traffic_vis}%
    }
    \\
    \subfloat[Visualization on Weather dataset]{%
        \begin{minipage}{0.31\textwidth}
            \includegraphics[width=\textwidth]{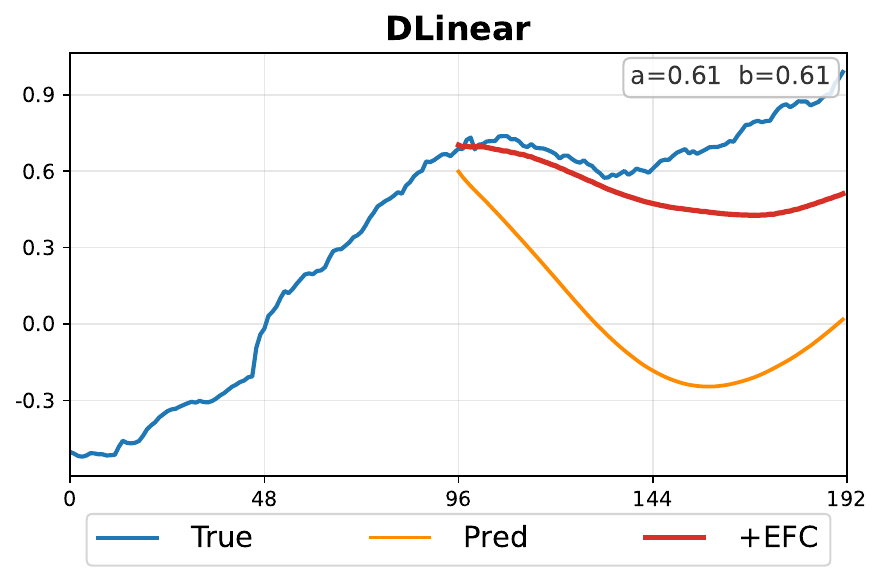}
        \end{minipage}%
        \hspace{-2pt}%
        \begin{minipage}{0.31\textwidth}
            \includegraphics[width=\textwidth]{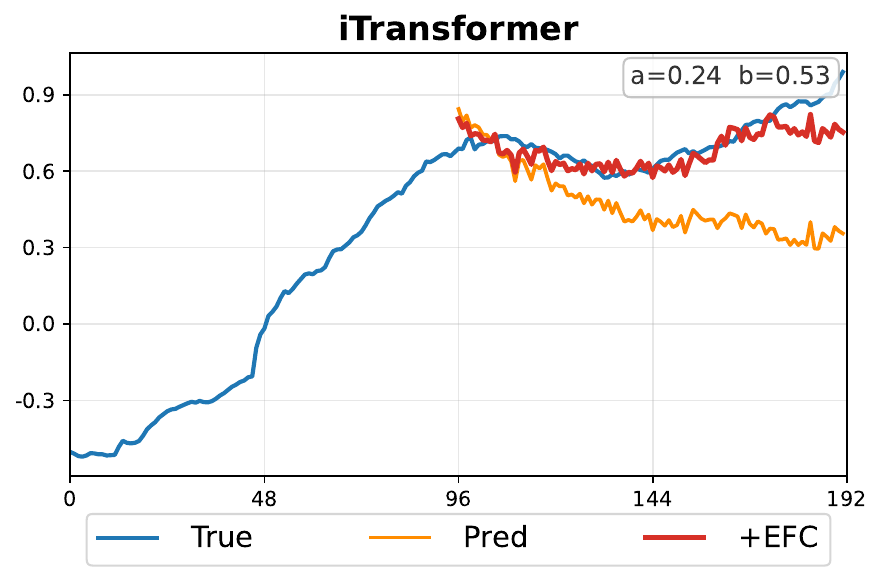}
        \end{minipage}%
        \hspace{-2pt}%
        \begin{minipage}{0.31\textwidth}
            \includegraphics[width=\textwidth]{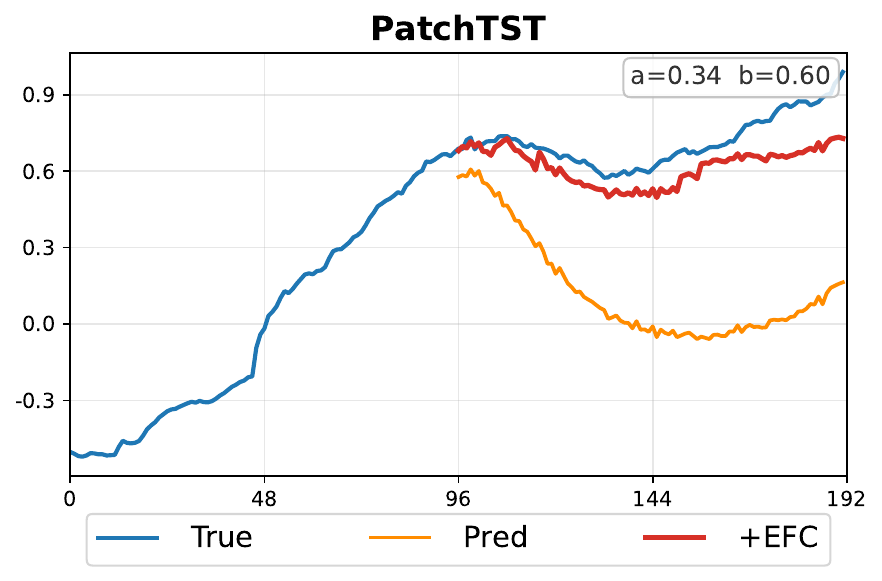}
        \end{minipage}%
        \label{fig:weather_vis}%
    }
    \caption{Forecasting visualization for backbone models with and without EFC. Representative forecasting samples are on Electricity, Traffic, and Weather datasets using DLinear, iTransformer, and PatchTST. Pred denotes the frozen backbone forecast, while +EFC denotes the corrected forecast.}
    \label{fig:forecast_visual}
\end{figure*}
\subsection{Hyperparameter Sensitivity Analysis}
We further investigate the sensitivity of EFC to four key hyperparameters, including the error history length $R$, input projection dimension $z$, EMA smoothing coefficient $\rho$, and GRU hidden dimension $d_{\mathrm{hid}}$. Experiments are conducted with iTransformer as the backbone on ECL, Weather, and Traffic datasets, while varying one hyperparameter at a time and keeping the others fixed. As shown in Fig. \ref{fig:sensitivity}, EFC maintains stable forecasting performance across different parameter configurations. For the error history length $R$, the performance remains consistent across prediction horizons, indicating that EFC does not rely on a specific amount of historical error information. Similar trends are observed for $z$ and $\rho$, where moderate variations only introduce marginal changes in MSE. The hidden dimension analysis further shows that increasing the correction module capacity beyond a certain level does not consistently improve performance, suggesting that the proposed lightweight correction mechanism is sufficient for capturing deployment-time error dynamics. Overall, these results demonstrate that EFC is robust to hyperparameter choices and can achieve reliable correction without extensive parameter tuning.

\subsection{Forecasting Visualization}
To provide a more intuitive and comprehensive view of the effect of EFC, we present forecasting visualizations comparing the predictions of backbone models with and without EFC across three representative datasets. As illustrated in Fig. \ref{fig:forecast_visual}, the examples exhibit different forecasting error patterns. On Electricity, the frozen forecasts largely follow the recurring fluctuations, while EFC adjusts local deviations in level and amplitude. On Traffic, forecasting errors are more pronounced around sharp peaks and their subsequent declines. EFC improves the alignment of several peak regions, although some local discrepancies remain. On Weather, the frozen predictions exhibit a sustained downward deviation from the observed trajectory, while EFC reduces this bias across the three backbones. These examples illustrate that error feedback can support different forms of forecast correction, ranging from local adjustments to more persistent bias correction. By using previously verified forecasting errors, EFC refines the predicted trajectories without modifying the underlying backbone.


\section{Conclusion}
In this work, we investigate verified forecasting errors as a source of feedback for correcting future predictions during deployment. Although modern forecasting models and foundation models provide strong predictive capabilities, their historical errors revealed during deployment still contain useful information for improving future predictions. To exploit this information without modifying the pretrained forecaster, we propose Error Feedback Calibration (EFC), a lightweight post-hoc correction framework that constructs causal error feedback signals, learns stable error transport, and models residual error dynamics through dedicated correction components. Extensive experiments on diverse real-world benchmarks demonstrate that EFC broadly improves forecasting accuracy while introducing limited computational overhead. In the future, we will investigate more general error feedback mechanisms for probabilistic forecasting, adaptive forecasting systems, and broader sequential prediction tasks.

\bibliographystyle{IEEEtran}
\bibliography{ref}

\end{document}